\documentclass[
]{ceurart}

\usepackage{listings}
\newcommand{\cconcept}[2]{\textcolor{#1}{\textit{#2}}}
\usepackage{xcolor}
\usepackage{multirow}
\usepackage{subcaption}

\usepackage{arydshln}
\usepackage{dashrule}

\definecolor{testimony}{RGB}{0,0,0}
\definecolor{evidence}{RGB}{0,0,0}
\definecolor{crime}{RGB}{0,0,0}
\definecolor{punishment}{RGB}{0,0,0}
\definecolor{motive}{RGB}{0,0,0}
\definecolor{process}{RGB}{0,0,0}
\usepackage{amsmath}
\usepackage{xcolor}
\usepackage{tikz}
\usepackage{tcolorbox}
\newcommand\mybox[2][]{\tikz[overlay]\node[fill=blue!20,inner sep=2pt, anchor=text, rectangle, rounded corners=1mm,#1] {#2};\phantom{#2}}
\newcommand{\gbox}[1]{\mybox[fill=green!20]{#1}}

\usepackage{arydshln}
\definecolor{testimony}{RGB}{0,0,0}
\definecolor{evidence}{RGB}{0,0,0}
\definecolor{crime}{RGB}{0,0,0}
\definecolor{punishment}{RGB}{0,0,0}
\definecolor{motive}{RGB}{0,0,0}
\definecolor{process}{RGB}{0,0,0}
\begin{document}


\conference{ArXiv 2026.}

\title{Can Causal Discovery Algorithms Help in Generating Legal Arguments?}


\author[1]{Soham Wasmatkar}[%
email=soham.2430010342@muj.manipal.edu,
]
\fnmark[1]
\address[1]{Manipal University Jaipur, Jaipur, Rajasthan, 303007, India}

\author[2]{Subinay Adhikary}[%
orcid=0009-0006-2165-2077,
email=sa21rs094@iiserkol.ac.in,
]
\fnmark[1]
\address[2]{Indian Institute of Science Education and Research (IISER) Kolkata, Mohanpur, 741246, India}

\author[2]{Rakshit Rohan}[%
orcid=0009-0000-7725-8550,
email=rr24rs102@iiserkol.ac.in,
]
\fnmark[1]

\author[3]{Shouvik Kumar Guha}[%
orcid=0000-0003-1623-0444,
email=shouvikkumarguha@nujs.edu,
url=https://www.nujs.edu/faculty/dr-shouvik-kumar-guha,
]
\address[3]{The West Bengal National University of Juridical Sciences, Kolkata, 700098, India}

\author[2]{Saptarshi Pyne}[%
orcid=0000-0001-9710-6749,
email=saptarshipyne01@gmail.com,
url=https://sap01.github.io,
]
\cormark[1]

\author[2]{Kripabandhu Ghosh}[%
orcid=0000-0002-8130-1221,
email=kripa.ghosh@gmail.com,
url=https://sites.google.com/view/kripabandhughosh-homepage,
]
\cormark[1]

\cortext[1]{Corresponding author.}
\fntext[1]{Joint first authors. They contributed equally.}

\begin{abstract}
  In 2011, Judea Pearl received the Turing Award, considered the Nobel Prize in Computing, for {\it fundamental contributions to artificial intelligence through the development of a calculus for probabilistic and causal reasoning.} It includes pioneering the development of {\it causal discovery} algorithms. These computer algorithms can analyze large multivariate datasets and automatically discover the causal relationships among the constituent variables. They have been widely used in many critical fields such as medicine and economics to support decisions. However, to our knowledge, they have not been leveraged in law. This paper attempts to alleviate this gap by
  investigating whether causal discovery algorithms can be leveraged for automated generation of legal arguments. To that end, a novel legal dataset is prepared by identifying 17 legal concepts, such as physical assault and property dispute. A curated collection of 150 homicide cases are annotated with these concepts, e.g., a case is annotated with physical assault only if a physical assault had been reported in that case. Subsequently, a selected set of widely-used causal discovery algorithms is applied to the annotated dataset to discover the causal relationships between the legal concepts. Additionally, the degrees of belief associated with the discovered relationships are quantified in mathematical probabilities. It is shown that some of the causal relationships help generate viable legal arguments, e.g., if one could establish that a physical assault has not taken place during a homicide, it should be a sufficient condition (with probability 1) to establish that the homicide has not been committed due to a property-related dispute. Thus, this paper shows that causal discovery algorithms can be helpful in generating legal arguments, opening up avenues for promising future endeavors. 
  Concurrently, a broader opportunity lies in investigating which other legal applications can benefit from the use of causal discovery algorithms.
\end{abstract}


\maketitle

\section{Introduction}

{\it Causal discovery} algorithms are computational methods for automating the discovery of causal relationships, among variables of interest, from large datasets \cite{Zanga2022CadisReview, Assaad2022CadisReview}. These algorithms are widely used in many critical fields, such as medicine, psychology, and economics, to support decisions \cite{Shen2020CadisMedicine, Anker2019CadisPsych, Addo2021CadisEcon}. However, their applications have not been leveraged in law, which is also a critical field. Instead, the legal literature only leverages the causal relationships discovered manually by legal experts \cite{Liepina2025CausalLaw, Dawid2025CausalLaw, Dahlman2022CausalLaw, Liefgreen2021CausalLaw, Liepina2020CausalLaw, Liepina2015CausalLaw}. 

\begin{figure}[h]
	\centering
	\includegraphics[width=0.7\textwidth]{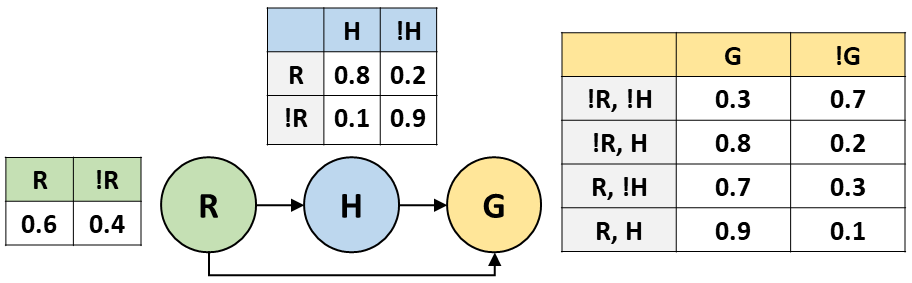}
	\caption{A Hypothetical Causal Graph between Three Variables--a Defendant being a Relative (R) of the victim, an heir (H) to the victim's wealth, and found to be guilty (G). Each of these nodes are assumed to be binary variables, e.g., a defendant can either be a relative (R) of the victim or not (!R). Each directed edge represents a cause-and-effect relationship between two nodes, e.g., the edge R $\rightarrow$ H indicates that being a relative has a causal effect on being an heir. In this paper, the degree of belief in the causal effect is measured as a ``conditional probability table'' (CPT). Each node is annotated with a CPT where the states of the node constitute the columns and the states of its cause(s), if any, constitute the rows. For example, in the CPT of node ``H'', the value 0.8 corresponding to row ``R'' and column ``H'' implies that, among the prior cases where the defendants were relatives of the victims, in 80\% of those cases, the defendants were also heirs.}
	\label{fig:causalGraph_RHG}
\end{figure}

Causal discovery algorithms can augment the invaluable insights of the legal experts by substantially enhancing the scale (i.e.\ the number of prior cases that can be analyzed) and quantifying the degree of belief for each causal relationship in a data-driven manner. For example, a causal discovery algorithm can be applied on a dataset that contains the files of all past homicide cases in the Supreme Court of India, the highest court of India. Suppose, the algorithm discovers a set of causal relationships between three variables
as depicted in Figure~\ref{fig:causalGraph_RHG}.

Subsequently, legal experts can examine whether the causal relationships align with their insights and make refinements, if necessary. The refined causal graph can be used for automatically generating quantitative arguments such as ``If you can establish that the defendant is not an heir to the victim's wealth, it increases the chance of proving the defendant not guilty by 59\%.'' The value of 59\% comes from calculating the conditional probabilities P(not guilty given not an heir) (denoted by $P(!G | !H)$ and P(not guilty given an heir) (denoted by $ P(!G | H)$, and computing their difference. $P(!G | !H)$ can be calculated as 
\begin{math}
	P(!G | !H) = P(!G, !H)/P(!H) \text{[By Bayes' theorem]}\\= \left(P(!G, R, !H) + P(!G, !R, !H)\right)/P(!H) \text{[By the law of total } \text{probability]}\\= P(!G | R, !H) * P(!H | R) + P(!G | !R, !H) * P(!H | !R) \text{[By Bayes' theorem]}\\ = 0.3 * 0.2 + 0.7 * 0.9 = 0.69
\end{math}.
Similarly, $P(!G | !H)$ calculates to $0.1$. Their difference is $\left(0.69 - 0.1\right) = 0.59$ which is the improvement in the probability if it can be established that the defendant is not an heir. Thus, the quantification of the arguments generated with the help of causal discovery algorithms are completely explainable in legal and mathematical terms. Moreover, the accuracy of the quantified values depends on the magnitude of prior cases a causal discovery algorithm has analyzed, which for a computer algorithm can be all related cases in human history.

The purpose of this paper is to employ some of the widely-used causal discovery algorithms on a moderately large number of legal cases and investigate whether the resultant causal graphs can help generate legal arguments. To that end, a novel dataset consisting of $150$ homicide cases is prepared in Section \ref{sec:meth}. Six widely-used causal discovery algorithms are applied on the dataset and the resultant causal graphs are examined to understand what kind of arguments can be generated from them (Section \ref{sec:results}). The conclusions are drawn in Section \ref{sec:conclusions} along with a discussion on the future directions.

\section{Methodology}
\label{sec:meth}
The preparation of the novel dataset and selection of the algorithms are described in this section.

\subsection{Dataset Preparation}

\begin{figure}[!h]
	\centering
	\includegraphics[width=1.12\linewidth]{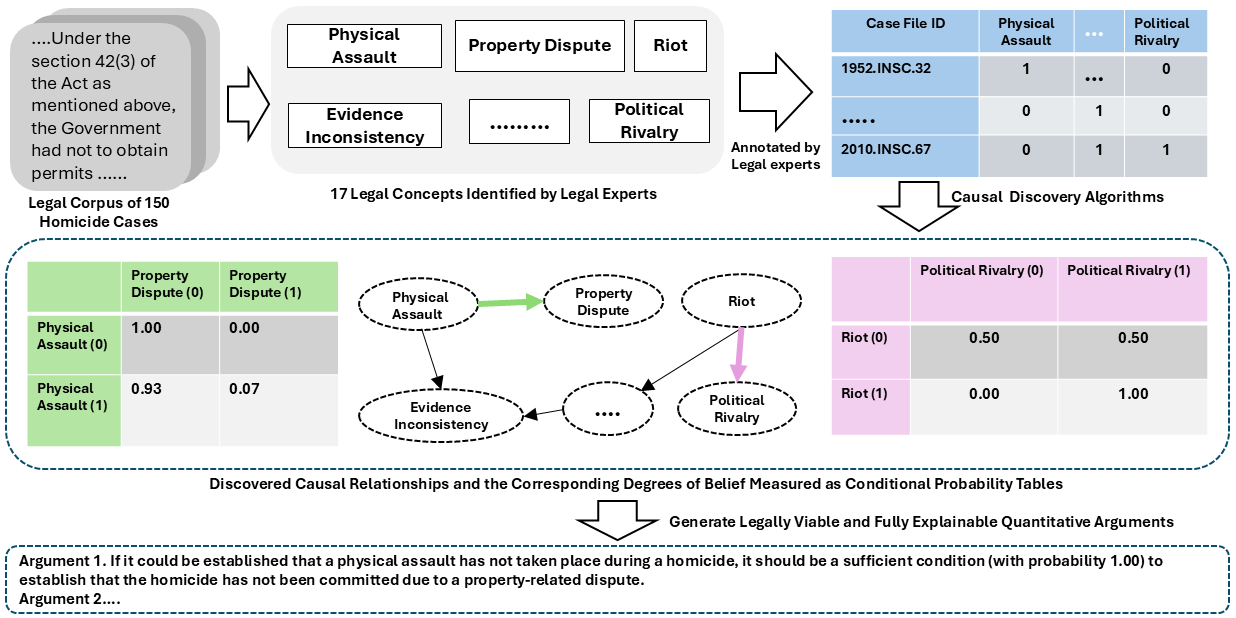}
	\caption{The Causal Discovery Algorithm-Aided Workflow for Automated Generation of Legal Arguments. A curated legal corpus of 150 homicide cases are utilized to identify 17 legal concepts. Subsequently, each case is annotated with a legal concept only if the concept is applicable to the case. Based on the presence (0) or absence (1) of the 17 legal concepts in 150 homicide cases, a 150-by-17 binary data matrix is prepared. The matrix is inputted to an array of widely-used causal discovery algorithms for discovering causal relationships among the legal concepts. The degrees of belief associated with the discovered causal relationships are measured as conditional probability tables (only two tables are shown). Finally, the causal relationships along with the associated conditional probability tables are used for generating legal arguments.}
	\label{fig:pipeline}
\end{figure}

\citet{adhikary2024case} identify seven legal concepts such as ``life imprisonment'' and ``physical assault'' (Table~\ref{tab:label_description}). Subsequently, they annotate $200$ homicide cases with these concepts, e.g., a case is annotated with the concept ``life imprisonment'' only if the court has punished the accused with life imprisonment. Ten new legal concepts are identified in this paper (Table~\ref{tab:label_description}). Among the aforementioned $200$ case files, $150$ are randomly selected for annotation. These case files are annotated with all of the (7+10) = 17 legal concepts by three legal experts using the legal annotation tool LeDA \cite{adhikary2023leda} (Figure~\ref{fig:example_annotation}).
The inter-annotator agreement (IAA) is measured using the F1-score, instead of the kappa coefficient, as recommended by \citet{wyner2013case}. The IAA score depends on three types of agreements: \textbf{Correct:} when two annotators mark exactly the same text span with the same concept, \textbf{Partial:} when both annotators assign the same concept but select different text spans, and \textbf{Missing:} when the annotators assign different labels for the same text span. For each case file, the IAA score is calculated with respect to each pair of annotators (Figure~\ref{fig:IAA_Computation}). The pairwise IAA scores are averaged to obtain the final IAA score of the case. The case files with IAA scores below 0.8 are reviewed by a senior legal expert who finalize the annotations.
Based on the annotations, we derived a binary data matrix encoding the presence (1) or absence (0) of the concepts in each case (Figure~\ref{fig:pipeline}), which is used as input to the causal discovery algorithms.

\begin{table*}[t]
	\centering
	\caption{The Legal Concepts Used for Annotating and Analyzing Homicide Cases. The concepts placed above the dashed line are identified by \citet{adhikary2024case} and the rest are identified by the authors of this paper.}
	\begin{tabular}{p{4cm} p{12cm}}
		\toprule
		\textbf{Legal Concept} & \textbf{Description} \\
		\midrule
		Life Imprisonment & Court punished the accused with life imprisonment for murder. \\
		
		Physical Assault & Accused killed, murdered, or seriously hurt a person using a sharp weapon or firearm. \\

		Homicide Murder & Court observed that the accused committed murder deliberately. \\
		
		Evidence Inconsistency & Proper or consistent evidence of the crime was not found in the case. \\
		
		Witness Testimony & Witness testimony was collected from non-expert people (e.g., a common person). \\
		
		Expert Witness Testimony & Witness testimony collected from forensic, ballistic, or other subject-matter experts. \\
		
		Riot & A riot was reported. \\
		
		\hdashrule{4cm}{1pt}{2pt} & \hdashrule{12cm}{1pt}{2pt}\\
		Political Rivalry & Incident was connected to political issues, resulting in injury or murder. \\
		Homicide Not Murder & Court declared the offense as culpable homicide not amounting to murder. \\

		
		Revenge & Murder committed as an act of vengeance. \\
		
		Property Dispute & Murder committed due to a property-related dispute. \\
		
		Testimony Challenged & Witness testimony presented by prosecution or defence was contested and evaluated by the court. \\
		
		Prosecutorial Delay or Inability & Prosecutorial delay or inability was recorded. \\
		
		Investigation Agency & Presence of investigative agencies other than the Police. \\
		
		Evidence Insufficient & Insufficiency of evidence was noted during the judgment. \\
		
		Rarest of the Rare Case & Court declared the case as a ``rarest of the rare'' case. \\
		
		Death Sentence & A death penalty was handed down by the court. \\
		\bottomrule
	\end{tabular}
	\label{tab:label_description}
\end{table*}

\begin{figure}[t]
\centering
\begin{tcolorbox}[
]
\scriptsize
 ...  The deceased Bheemanna, who was present on his land along with his wife Paddamma (PW.1) and his mother Bheemava, obstructed the accused persons and asked them not to pass through his land. Yenkappa (A-1) then began hurling abuses in filthy language and instigated
$\underbrace{\text{\gbox{Bhimanna (A-2), Suganna (A-3), and other accused persons, forming an unlawful assembly,}}}_{Concept: \text{\textbf{Riot}}}$  to collectively engage in violence against the deceased.  Thereafter, the said accused persons, 
$\underbrace{\text{\gbox{acting in furtherance of the unlawful assembly, assaulted the deceased with axes on his head and right hand.}}}_{Concept: \text{\textbf{Riot, Homicide Murder}}}$ 
... .... Upon the advice of the police, the deceased was taken in a mini lorry, driven by Mahadevappa (PW.10) to Deodurga Hospital, and when they reached there at 8.00 p.m., $\underbrace{\text{\gbox{based on the complaint submitted by Paddamma (PW.1), an FIR was lodged at 8.15 p.m.}}}_{Concept: \text{\textbf{ Homicide Murder}}}$
\end{tcolorbox}
\caption{
Examples of How Text Spans Within or Across Sentences are Annotated with Legal Concepts. 
}
\label{fig:example_annotation}
\end{figure}


\begin{figure}[!h]
    \centering
  \begin{tcolorbox}[
]
\scriptsize


\textbf{Annotator 1:}
\begin{itemize}
    \item \textit{``Bhimanna (A-2), Suganna (A-3), and other accused persons, forming an unlawful assembly''}
    $\rightarrow$ \textbf{Riot}
    
    \item \textit{``assaulted the deceased with axes on his head and right hand.''}
    $\rightarrow$ \textbf{Homicide Murder, Riot}
    
    \item \textit{``an FIR was lodged at 8.15 p.m.''}
    $\rightarrow$ \textbf{Homicide Murder}
\end{itemize}

\textbf{Annotator 2:}
\begin{itemize}
    \item \textit{``Bhimanna (A-2), Suganna (A-3), and other accused persons, forming an unlawful assembly''}
    $\rightarrow$ \textbf{Riot}
    
    \item \textit{``assaulted the deceased with axes on his head and right hand''}
    $\rightarrow$ \textbf{Homicide Murder, Riot}
    
    \item \textit{``complaint submitted by Paddamma (PW.1), an FIR was lodged''}
    $\rightarrow$ \textbf{Homicide Murder}
\end{itemize}

\vspace{2pt}
\textbf{Agreement counts:}
\[
\textcolor{green!50!black}{Correct (C) =1} \quad
\textcolor{orange!80!black}{Partial (P) =2} \quad
\textcolor{red!60!black}{Missing (M) =0} \quad
\]

\textbf{Weighted scoring} (partial weight $=0.5$):
\[
\text{Precision} = \text{Recall} = \frac{C + 0.5P}{C + M + P} \approx 0.67; \quad
\textbf{IAA Score (F1)} = 
\frac{(1+\beta^2)\cdot \text{Precision}\cdot \text{Recall}}
{\beta^2 \cdot \text{Precision} + \text{Recall}} = \mathbf{0.67} 
\quad ,\hspace{0.5em} 
\]
{where $\beta$ = 1, giving an equal weight to precision and recall.}

\end{tcolorbox}
    \caption{
    An Example of How the Inter-Annotator Agreement (IAA) Score is Calculated.
    }
    \label{fig:IAA_Computation}
\end{figure}

\subsection{Selection of Algorithms}
Six widely-used causal discovery algorithms, namely, PC~\cite{spirtes2000causation}, GES \cite{chickering2002optimal}, GRaSP~\cite{lam2022greedy}, BOSS~\cite{BOSS}, LiNGAM~\cite{shimizu2006linear}, and ANM~\cite{hoyer2008nonlinear}, are selected based on their methodological diversity. PC constructs the causal graph by adding one causal relationship at a time. It does so by performing a ``conditional independence'' test on each potential relationship and adding the relationship only if it passes the test. GES, on the other hand, proactively adds a relationship and examines whether this addition increases the likelihood-based ``score'' of the whole graph. The relationship is removed only if it does not increase the score. GRaSP and BOSS start with the variables (here, legal concepts) arranged in a random order (suppose, ``R H G'' for the Figure~\ref{fig:causalGraph_RHG} example). They construct all causal graphs that can be represented by this order s.t.\ for every causal relationship $X \rightarrow Y$ in the causal graph, $X$ should appear before $Y$ in the given order, e.g., the causal graph in Figure~\ref{fig:causalGraph_RHG} is one of the valid causal graphs for the order ``R H G''. The algorithms calculate different permutations of the given order (e.g., ``H R G''), calculates the scores of all causal graphs for each permuted order, and selects a causal graph with the maximal score. LiNGAM and ANM formulate the causal discovery task as linear and non-linear regression problems, respectively, where a legal concept can be modeled as the target variable and its potential causes i.e.\ the remaining legal concepts can be modeled as predictor variables.

\section{Results}
\label{sec:results}
\begin{figure}[h]
	\centering
	\includegraphics[width=0.6\linewidth]{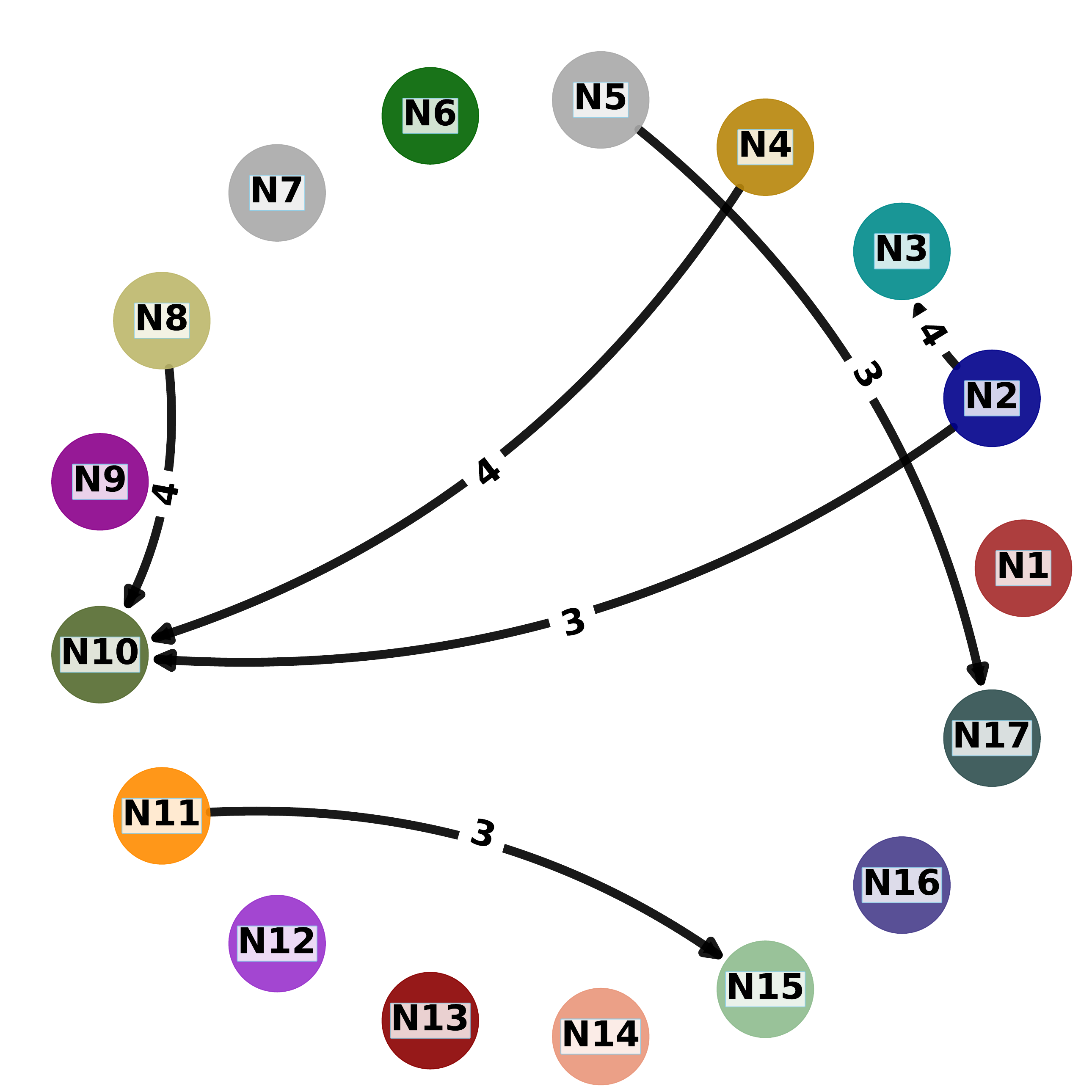}
	\caption{The Consensus Causal Graph. The nodes represent the legal concepts. 
		\cconcept{testimony}{N1}: witness testimony,
		\cconcept{process}{N2}: prosecutorial delay or inability,
		\cconcept{testimony}{N3}: testimony challenged,
		\cconcept{crime}{N4}: riot,
		\cconcept{punishment}{N5}: death sentence,
		\cconcept{punishment}{N6}: life imprisonment,
		\cconcept{crime}{N7}: homicide murder,
		\cconcept{evidence}{N8}: evidence inconsistency,
		\cconcept{testimony}{N9}: expert witness testimony,
		\cconcept{motive}{N10}: political rivalry,
		\cconcept{crime}{N11}: physical assault,
		\cconcept{evidence}{N12}: evidence insufficient,
		\cconcept{process}{N13}: investigation agency,
		\cconcept{motive}{N14}: revenge,
		\cconcept{motive}{N15}: property dispute,
		\cconcept{crime}{N16}: homicide not murder,
		and
		\cconcept{punishment}{N17}: rarest of the rare case.
		The edges represent the causal relationships agreed upon by multiple algorithms. An edge weight (e.g., 3 or 4) implies the count of algorithms that agree on the edge. 
	}
	\label{fig:consensus_graph}
\end{figure}

Each of the six causal discovery algorithms outputs a causal graph that represents the causal relationships discovered by that algorithm. To figure out which causal relationships are true causal patterns hidden in the dataset and not sensitive to the choice of an algorithm, a ``consensus causal graph'' is constructed (Figure~\ref{fig:consensus_graph}). It consists of the causal relationships that are independently discovered by more than one algorithm. Each relationship is assigned an integer weight (e.g., 3) which represents how many algorithms agree on the relationship.

There are six causal relationships in the consensus causal graph and all of them are agreed on by half or more of the algorithms, indicating how robust the relationships are. They cater to four ``effect'' nodes: political rivalry (3 causal relationships) and testimony challenged, rarest of the rare case, property dispute (1 causal relationship each). The degrees of belief in their discovered causes are measured in CPTs. Some of them are examined below.

\paragraph{\textbf{The Causal Relationships of Political Rivalry}}
Political rivalry has three ``predictive causes"--evidence inconsistency, prosecutorial delay or inability, and riot (see node N10 in Figure \ref{fig:consensus_graph}). They are called predictive causes since the combinations of their values can predict what value political rivalry would take \cite{Diebold1998PredCausality}. These are different from the true cause of political rivalry, which involves sociopolitical history. Three algorithms, namely, PC, GRaSP, and BOSS, agree on all three of these causes (Table \ref{tab:pol_riv_3c}) while the GES algorithm agrees on only two of these causes (Table \ref{tab:pol_riv_2c}). From Table \ref{tab:pol_riv_3c}, it can be seen that, for all the prior cases where consistent evidence was not found (N8 = 1) and riots were reported (N4 = 1), the homicides were connected to political rivalry. In other words, if consistent evidence is not found and a riot is reported in a homicide case, it can be argued that the homicide(s) is connected to political rivalry with certainty (probability = 1.0). For all other combinations of their values, the connection to political rivalry is either non-existent (probability = 0.0) or equivalent to a random chance (probability = 0.5). Although, the value of prosecutorial delay or inability (N2) is irrelevant for reaching a certainty, it plays a crucial role in increasing the probability from non-existence to a random chance. GES, being a local optimization algorithm, could not capture the role of prosecutorial delay or inability in determining political rivalry. However, it does agree on the roles of evidence incosistency and riot in ascertaining political rivalry (Table \ref{tab:pol_riv_2c}). 

The proposed framework is substantially different from those based on pairwise correlation measures, such as the Pearson correlation coefficient \cite{Lee1988Corr}. Firstly, the causal relationships are directed. Secondly, it examines a fine-grained association between a distinct value-combination of potential causes and a distinct value of the effect. Even though there are only two cases involving political rivalry (Column ``N10 = 1", Table~\ref{tab:pol_riv_3c}) in the dataset, it could detect that evidence inconsistencies and riots are simultaneously reported in exactly those cases. Pearson correlation, on the other hand, measures the association between political rivalry and a single potential cause, which aggregates over their different combinations of values. Since there are much larger number of cases involving evidence inconsistencies (12+1+0+1 = 14 cases) and riots (9+1+1+1 = 12 cases), their Pearson correlations with political rivalry are reported to be low, 0.36 and 0.39, respectively. Hence, these predictive causes could have never been discovered using such correlation measures.

\begin{table}[h]
	\caption{The Conditional Probability Table of ``Political Rivalry'' in the Causal Graphs Separately Discovered by the PC, GRaSP, and BOSS Algorithms. The rows where the causes predict the presence of political rivalry (N10 = 1) with certainty i.e.\ with a probability of 1.0 are boldfaced. The number of cases corresponding to each probability value is given in parentheses.}
	\begin{tabular}{llll}
		\toprule
		\multicolumn{1}{c}{\textbf{Causes}} & \multicolumn{1}{c}{\textbf{Effect}} \\ \midrule
		\textbf{\begin{tabular}[c]{@{}l@{}}Evidence Inconsistency (N8),\\ Prosecutorial Delay or Inability (N2), \\ Riot (N4)\end{tabular}} & \textbf{\begin{tabular}[c]{@{}l@{}}Political\\ Rivalry \\ (N10) = 0\end{tabular}} & \textbf{N10 = 1} & \textbf{\begin{tabular}[c]{@{}l@{}}Total No. of Cases for Each \\Combination of N8,N2,N4\end{tabular}} \\ \midrule
		0, 0, 0 & 1.0 (125) & 0.0 (0) & (125) \\ \hline
		0, 0, 1 & 1.0 (9) & 0.0 (0) & (9) \\ \hline
		0, 1, 0 & 1.0 (1) & 0.0 (0) & (1) \\ \hline
		0, 1, 1 & 1.0 (1) & 0.0 (0) & (1) \\ \hline
		1, 0, 0 & 1.0 (12) & 0.0 (0) & (12) \\ \hline
		1, 0, 1 & 0.0 (0) & \textbf{1.0} (1) & (1) \\ \hline
		1, 1, 0 & 0.5 (0) & 0.5 (0) & (0) \\ \hline
		1, 1, 1 & 0.0 (0) & \textbf{1.0} (1) & (1) \\ \midrule
        Total no. of cases for each value of N10 & (148) & (2) & (150) \\ 
        \bottomrule
	\end{tabular}
	\label{tab:pol_riv_3c}
\end{table}

\begin{table}[h]
	\caption{The Conditional Probability Table of ``Political Rivalry'' in the Causal Graph Discovered by the GES Algorithm. The row where the causes predict the presence of political rivalry (N10 = 1) with certainty (probability = 1.0) is boldfaced.}
	\begin{tabular}{lll}
		\toprule
		\multicolumn{1}{c}{\textbf{Causes}} & \multicolumn{2}{c}{\textbf{Effect}} \\ \hline
		\textbf{\begin{tabular}[c]{@{}l@{}}Evidence Inconsistency (N8),\\ Riot (N4)\end{tabular}} & \textbf{\begin{tabular}[c]{@{}l@{}}Political\\ Rivalry \\ (N10) = 0\end{tabular}} & \textbf{N10 = 1} \\ \hline
		0, 0 & 1.0 & 0.0 \\ \hline
		0, 1 & 1.0 & 0.0 \\ \hline
		1, 0 & 1.0 & 0.0 \\ \hline
		1, 1 & 0.0 & \textbf{1.0} \\ \bottomrule
	\end{tabular}
	\label{tab:pol_riv_2c}
\end{table}

\paragraph{\textbf{The Causal Relationship between Physical Assault and Property Dispute}}
Half of the causal discovery algorithms agree that there exists a causal relationship between physical assaults (N11) and motives related to property disputes (N15) in homicide cases (Figure \ref{fig:consensus_graph}). They concur that establishing a physical assault has not taken place during a homicide should be a sufficient condition (with probability = 1.00) to establish that the homicide has not been committed due to a property-related dispute (Table \ref{tab:pro_dis}). Such a strong causation could not have been captured from their extremely low Pearson correlation (0.2). On the other hand, establishing that a physical assault has taken place does not suffice to establish that the homicide is related to a property dispute. It only increases the chance by 7\% (from a probability value of 0.00 to 0.07). The causal factors that can explain the remaining 93\% of uncertainty in establishing a motive related to property dispute might be beyond the seventeen legal concepts considered in this paper. Identifying more legal concepts and annotating prior cases with them can help causal discovery algorithms mitigate this gap. The dataset and code used in this section can be found at \textbf{\url{https://github.com/spynegroup/LegalCausation}}.

\begin{table}[h]
	\caption{The Conditional Probability Table of ``Property Dispute'' in the Consensus Causal Graph (Figure \ref{fig:consensus_graph})}
	\begin{tabular}{llll}
		\toprule
		\multicolumn{1}{c}{\textbf{Cause}} & \multicolumn{2}{c}{\textbf{Effect}} \\ \hline
		\textbf{\begin{tabular}[c]{@{}l@{}}Physical Assault (N11)\end{tabular}} & \textbf{\begin{tabular}[c]{@{}l@{}}Property Dispute\\(N15) = 0\end{tabular}} & \textbf{N15 = 1} & \textbf{\begin{tabular}[c]{@{}l@{}}Total No. of Cases for\\Each Value of N11\end{tabular}} \\ \hline
		0 & 1.00 (80) & 0.00 (0) & (80) \\ \hline
		1 & 0.93 (65) & 0.07 (5) & (70) \\
        \midrule
        Total no. of cases for each value of N15 & (145) & (5) & (150) \\
        \bottomrule
	\end{tabular}
	\label{tab:pro_dis}
\end{table}

\section{Conclusions and Future Work}
\label{sec:conclusions}
This paper asks the question, ``Can causal discovery algorithms help in generating legal arguments?''.\ To answer the question, a novel legal dataset is prepared by annotating 150 prior homicide cases with 17 legal concepts such as the \textit{life imprisonment}. A case is annotated with \textit{life imprisonment} only if the court imposed life imprisonment on the accused in that case. The annotated case files are separately inputted to six widely-used causal discovery algorithms. Each algorithm discovers a distinct set of causal relationships. To identify robust causal relationships, only those that are discovered by more than one algorithm are considered for further analysis. Many of these causal relationships are found to be useful for generating quantitative arguments that are legally viable and fully explainable, e.g., a consensus is achieved by half of the algorithms that establishing a physical assault has not taken place during a homicide should be a sufficient condition (with probability 1) to establish that the homicide has not been committed due to a property-related dispute.\\

Therefore, future directions lie in answering the question ``How can we maximize the utilization of causal discovery algorithms for automated legal argument generation?''.\ Increasing the number of legal concepts and annotating thousands of cases with them can be a mammoth task; however, it can enhance the quality and quantity of legal arguments produced with the help of causal discovery algorithms. At the same time, more advanced levels of causal reasoning techniques can be applied on the discovered causal relationships to generate more nuanced and persuasive legal arguments. The causal reasoning technique applied in this paper is known as \textit{association} based reasoning which forms the lowest level in \textit{the ladder of causation} \cite{Pearl2018Why}. The highest level is formed by the \textit{counterfactual} reasoning. A counterfactual is a scenario that did not happen but could have happened. For example, if person ``X'' keeps a couch in front of a fire exit and person ``Y'' dies in a fire, is X legally responsible for Y's death? It necessitates answering the causal question of whether the couch blocking the fire exit is a ``but-for'' cause of Y's death \cite{Spellman2001ButFor}; the answer lies in calculating the counterfactual probability that Y would have survived if the couch had not been blocking the fire exit. If the probability is reasonably high, one could argue that X should be held legally responsible for Y's death. In 2011, Judea Pearl received the \textit{Turing Award}, considered the Nobel Prize in Computing, for ``fundamental contributions to artificial intelligence through the development of a calculus for probabilistic and causal reasoning'' \cite{TuringAward2011} which includes the development of a theoretical framework for counterfactual reasoning in conjunction with causal discovery. Pearl is profoundly optimistic about its applications in law and believes ``In principle, counterfactuals should find easy application in the courtroom'' \cite{Pearl2018Why}.\\


For accused persons, witnesses, and victims, causal reasoning provides transparency about how decisions affecting their liberty and interests are made. For victims' families in death sentence cases, understanding the causal structure may help them evaluate whether the sentence reflects the crime's severity or whether procedural factors such as prosecutorial inability were determinative.\\

The causal discovery algorithms can be used to automatically discover a causal structure from prior related cases in a fast and fully transparent manner. This paper envisages a pipeline where the automatically discovered causal structure is refined through the human supervision of legal experts. The refined causal structure can help the litigating parties understand the causal reasoning behind the judgment affecting their lives and , if required, appeal if the said causal reasoning diverges from the refined causal structure. Moreover, many causal discovery algorithms can incorporate domain-expert knowledge to improve efficiency and reduce false discovery \cite{Yu2024MPPC}. This feature can be utilized to incorporate causal relationships known a priori such as the ones specified by legal doctrines. One may therefore opine that incorporating the usage of causal discovery algorithms in legal analysis represents not merely a technical upgrade, but a conceptual advancement toward law's fundamental commitment to timely justice and transparency.

\section*{Declaration on Generative AI}
The authors have not employed any Generative AI tools.
  

\bibliography{main}

\appendix

%
%

\end{document}